\documentclass{article}
\usepackage{spconf,amsmath,amssymb,graphicx}
\usepackage{color}
\usepackage{cite}
\usepackage{balance}

\usepackage{wasysym}
\usepackage{bm}
\usepackage{booktabs}
\usepackage{etoolbox,siunitx}
\robustify\bfseries
\usepackage{caption}
\usepackage{enumitem} %
\usepackage{listings}
\usepackage{pifont}%
\usepackage[T1]{fontenc}
\usepackage{algpseudocode}
\usepackage{algorithm}
\makeatletter
\algnewcommand{\LineComment}[1]{\Statex \hskip\ALG@thistlm \(\triangleright\) #1}
\algnewcommand{\IndentLineComment}[1]{\Statex \hskip\ALG@tlm \(\triangleright\) #1}

\def\LL{{\mathcal{L}}}
\def\UU{{\mathcal{U}}}
\def\PP{{\mathcal{P}}}
\def\TT{{\mathcal{T}}}
\def\EE{{\mathcal{E}}}
\def\AA{{\mathcal{A}}}

\title{Unsupervised Domain Adaptation for Speech Recognition\\via Uncertainty Driven Self-Training}
\name{Sameer Khurana$^{1,2}$, Niko Moritz$^{1}$, Takaaki Hori$^{1}$, Jonathan Le Roux$^{1}$\thanks{This work was performed while S.~Khurana was an intern at MERL.}}
\address{$^1$Mitsubishi Electric Research Laboratories (MERL), Cambridge, MA, USA\\$^{2}$Massachusetts Institute of Technology, Cambridge, MA, USA } %

\begin{document}
\ninept
\maketitle
\begin{abstract}
The performance of automatic speech recognition (ASR) systems typically degrades significantly when the training and test data domains are mismatched. In this paper, we show that self-training (ST) combined with an uncertainty-based pseudo-label filtering approach can be effectively used for domain adaptation. We propose DUST, a dropout-based uncertainty-driven self-training technique which uses agreement between multiple predictions of an ASR system obtained for different dropout settings to measure the model's uncertainty about its prediction. DUST excludes pseudo-labeled data with high uncertainties from the training, which leads to substantially improved ASR results compared to ST without filtering, and accelerates the training time due to a reduced training data set.
Domain adaptation experiments using WSJ as a source domain and TED-LIUM 3 as well as SWITCHBOARD as the target domains show that up to 80\% of the performance of a system trained on ground-truth data can be recovered.

\end{abstract}
\section{Introduction} %
\label{sec:intro}
Over the past years, the performance of end-to-end automatic speech recognition (ASR) systems has improved dramatically. This success is driven by improved neural network architectures and training frameworks \cite{Graves2006, Graves2013, Chorowski2015, PoveyPGG16, Hori2017}, increasingly large amounts of labeled data \cite{librispeech, commonvoice}, and increased computational resources for training complex models. However, ASR performance  degrades significantly when the target domain (testing conditions) does not match the source domain (training data). Domain mismatch between training and testing conditions occurs commonly when ASR systems are deployed in the real world, with several factors contributing to it, such as dialectal and accent variations, speaking style (e.g., conversational vs read), and difference in acoustic conditions (e.g., noisy vs clean).
A straightforward approach to remedy this problem is to collect labeled data in the target domain and use it for adapting a pre-trained source model. However, manually annotating large amounts of data for every new target domain is expensive and time consuming. Thus, there is a need for unsupervised adaptation algorithms that can leverage unlabeled data for source to target domain adaptation \cite{bell2020adaptation}.

Recently, distribution alignment methods that do not require access to parallel data have become popular for unsupervised domain adaptation. These methods attempt to align the source and target data distributions. Some alignment tools that have shown promise are optimal transport \cite{DBLP:journals/corr/CourtyFTR15}, domain adversarial training with gradient reversal layer (GRL) \cite{ganin2016domain}, and training using discrepancy losses \cite{saito2018maximum}. Domain adversarial learning with GRL is used for ASR in \cite{adams2019massively, sun2018domain}. 

We focus on self-training (ST) \cite{scudder1965probability} for unsupervised domain adaptation. %
ST proceeds by training a teacher model on the labeled source domain data, which is used to generate pseudo-labels for the unlabeled target domain data to obtain pseudo-parallel data. A student model is then trained on the augmented training data including both labeled and pseudo-parallel data to obtain a model that is expected to generalize better to the target domain.
ST has recently shown excellent performance for neural sequence generation tasks such as machine translation \cite{he2019revisiting} and ASR \cite{hsu2020lpriormatch, weninger2020semisupervised, moritz2020gtc}, achieving state-of-the-art performance for semi-supervised ASR when applied in an iterative manner \cite{xu2020iterative}.
Classical works in ST \cite{nigam2000text, blum2001learning, 1512038} suggest that its performance is not stable if the generated pseudo-labels are highly erroneous, and hence ST is often accompanied by a filtering process to remove such pseudo-labeled utterances from the training data. %
However, recent work on ST has shown strong results with no filtering at all \cite{xu2020iterative}. We hypothesize that this is due to two key assumptions made in that work:
1) No mismatch between the source and target domain. Hence, the teacher model that is trained with labeled source domain data is able to generate relatively clean pseudo-labels for the unlabeled target domain data.
2) Access to large amounts of in-domain text data, which is used to build a strong language model (LM) that is applied for beam search decoding to generate the pseudo-labels for ST.

In this work, we consider the case where these two assumptions do not hold: we focus on a domain mismatch between the source and target data sets with access to ground-truth labels for the source domain only. %
In this case, the pseudo-labels generated by the teacher model for the unlabeled target domain data may be less accurate, which increases the need to apply a pseudo-label filtering strategy. To that end, we propose dropout-based uncertainty-driven self-training (DUST), which filters pseudo-labeled data based on the model's uncertainty about its prediction as measured using the degree of agreement between multiple transcriptions obtained with various realizations of dropout and a reference transcription obtained without dropout \cite{gal2016dropout,vyas2019analyzing}. %
We show that DUST %
is an effective method for mismatched domain adaptation and substantially improves over the baseline model, which is trained on the source domain labeled data only, as well as over iterative ST without filtering \cite{xu2020iterative}, 
whereby the largest gain is observed when the source and target domain mismatch is most severe. 
In addition, DUST leads to a faster and more efficient training compared to iterative ST, since the filtering process selects only a fraction of the whole unlabeled data set with reliable pseudo-labels.
Finally, we perform a preliminary study showing that DUST can be combined with a self-supervised representation learning approach for low-resource conditions.

\section{DUST}
\textbf{Using dropout to measure model uncertainty:}
DUST uses %
the model's uncertainty about its predictions $\hat{y}_u$ for an unlabeled target data point $x_u$ to weed out the pseudo-labeled pair $\{x_u, \hat{y}_u\}$, if the model's uncertainty is high.
Assuming the model involves dropout layers \cite{Srivastava2014}, uncertainty can be quantified by sampling multiple predictions from the model using dropout and computing agreement between the sampled predictions and a reference prediction obtained without dropout, with low agreement corresponding to high uncertainty. Intuitively, this filtering process can be understood as polling multiple experts for prediction about an unlabeled data point. If the predictions of all the experts agree on a particular data point, it is likely to be correct. Formally, the method can be understood using the work of Gal and Ghahramani \cite{gal2016dropout}, which made connections between Bayesian probability theory and neural networks trained with dropout. In particular, they show that a model's predictive variance approximately equals the sample variance of multiple %
stochastic passes through the network. Here, a stochastic pass refers to inference with a dropout realization. This technique is closely related to \cite{vyas2019analyzing}, which uses a model's prediction uncertainty computed using dropout to estimate word error rates.
DUST combines ST and pseudo-label filtering based on the ASR model's uncertainty for an unlabeled speech utterance using dropout.

\textbf{Self-training with DUST:} The overall DUST procedure is summarized in Algorithm 1. We assume that we have access to a set of labeled parallel data $\LL = \{x_i, y_i\}_{i=1}^{L}$ in a source domain, 
and a set of unlabeled data $\UU = \{x_j\}_{j=L+1}^{L+U}$ in a target domain,
with potentially a strong mismatch between the two domains. %
DUST proceeds by first training a base model $f^{p}_\theta$ on the labeled data $\LL$ with dropout layers, using a dropout probability $p\in[0,1]$. This base model is then used to provide predictions on the unlabeled data  $\UU$ to generate pseudo-parallel data, of which only a 
subset $\PP$ is selected based on the model's uncertainty on each unlabeled data point, as described further below.
Once the subset $\PP$ has been determined, a new model is trained on the labeled data $\LL$ augmented with the subset $\PP$ of pseudo-parallel data, and the procedure can be reiterated, with the newly trained model used as base model.

An unlabeled data point $x_u$ is considered for inclusion in the subset $\PP$ of selected pseudo-parallel data as follows. 
1) First, a reference hypothesis $\hat{y}_u^{\text{ref}}$ for $x_u$ is generated using the model with disabled dropout layers, resulting in a deterministic inference process which we refer to as deterministic forward pass. %
2) Second, multiple hypotheses $\hat{y}_u^t$ are sampled from the model by running it $T$ times with dropout using different random seeds in $\TT$, a process we refer to as stochastic forward pass. %
3) Finally, the Levenshtein edit distance %
between each of the $T$ sampled hypotheses and the reference hypothesis is computed, leading to a set $\EE$ of $T$ distances. The edit distance is normalized by the length of the reference hypothesis. If all the values in $\EE$ are below a pre-defined threshold ratio $\tau$ of the length $|\hat{y}_u^{\text{ref}}|$ of the reference hypothesis, then we add the pseudo-labeled data point $\{x_u, \hat{y}_u^{\text{ref}}\}$ to $\PP$,  otherwise we reject it. By setting the filtering threshold low, we can accept only pseudo-labeled data points on which stochastic samples have high agreement, which implies low sample variance and in turn implies low model predictive uncertainty \cite{gal2016dropout}. %
Our working hypothesis is that data points on which the model has a low predictive uncertainty should be good enough for self-training. We empirically show that low thresholds weed out the noisy pseudo-labels, i.e., inaccurate pseudo-labels (See Section~\ref{sec:results}). In practice, running beam search multiple times is computationally expensive and hence, we only run stochastic beam search $T=3$ times to draw three samples from the model.

\begin{algorithm}[t]
\label{algo:1}
\caption{Dropout-based Uncertainty-driven Self-Training (DUST)}
\begin{algorithmic}[1]
\State Given labeled data $\LL$ and unlabeled data $\UU$
\State Given a set $\TT$ that contains $T$ natural numbers %
\State Train a base model $f^{p}_\theta$, with dropout $p$, on labeled data $\LL$ %
\Repeat
\State Let $\PP$ be the set of selected pseudo-labeled data points
\State Let $\EE$ be a set of edit distances
\State Initialize $\PP$ and $\EE$ as empty sets
\ForAll{$x_u \in \UU$}
\State Compute deterministic forward pass $f^{0}_{\theta}(x_u)$
\State $\hat{y}^{\text{ref}}_u = \text{beam\_search}(f^{0}_\theta(x_u))$ %
\ForAll{$t \in \TT$}
\State Set random seed to $t$
\State Compute stochastic forward pass $f^{p}_{\theta}(x_u)$
\State $\hat{y}^t_u = \text{beam\_search}(f^{p}_\theta(x_u))$ %
\State $e = \text{edit\_distance}(\hat{y}_u^t, \hat{y}_u^{\text{ref}})$
\State Add $e$ to the set $\EE$
\EndFor
\If{$\text{max}(\EE) < \tau |\hat{y}_u^{\text{ref}}|$ (with $\tau$ a filtering threshold)} %
\State Add $\{x_u, \hat{y}_u^{\text{ref}}\}$ to the set $\PP$
\EndIf
\EndFor
\State Train a new model $f^{p}_\theta$ on $\AA=\LL \cup \PP$
\Until{convergence or maximum self-training iterations reached}
\end{algorithmic}
\end{algorithm}

\section{Experimental Setup}
\textbf{Data}: 
We use the Wall Street Journal \cite{paul1992design} (WSJ) dataset as our source domain, and TED-LIUM 3 \cite{hernandez2018ted} (TED) as well as Switchboard \cite{godfrey1992switchboard} (SWBD) as our target domains. WSJ is a clean read English news speech corpus consisting of 80 hours of labeled training data spoken by 280 speakers from different parts of the United States. TED consists of 450 hours of transcribed English Ted Talks on a wide range of topics by 2,000 speakers %
from all over the world. SWBD consists of 260 hours of two-sided telephone conversations among 543 speakers (302 male, 241 female) from all areas of the United States. The domain mismatch between source and target domains is quite evident and is most severe with SWBD as target. %

\textbf{Neural Network Architecture for ASR:} %
The neural network model consists of two functions, EncPre($\cdot$), which takes in the input speech sequence and outputs a sub-sampled sequence, and EncBody($\cdot$), which processes the subsampled sequence and outputs the logits for classification \cite{karita2019comparative}. The input speech is represented as a sequence of 80 dimensional log-mel spectral energies plus 3 pitch features. EncPre($\cdot$) is a 2 layer CNN with 256 channels, stride 2, and kernel size $3 \times 3$. EncBody($\cdot$) consists of 12 transformer blocks. Each block consists of a self-attention layer followed by two fully connected layers with an interleaved ReLU non-linearity. A dropout layer is applied after self-attention and each fully connected layer. Layer-Norm is used after both self-attention and the two fully connected layers. The number of neurons in the first fully connected layers is 1024. Each self-attention layer consists of 4 attention heads with attention vector dimension of 256. We set the dropout rate to $0.1$ during training and use the same dropout rate when sampling predictions from the model for filtering.

\textbf{Data Augmentation:} We apply two different data augmentation methods:
1) an offline data augmentation approach that uses simulated room impulse responses and point-source noises to augment clean speech utterances \cite{ko2017study}, and
2) SpecAugment \cite{park2019specaugment}, %
where we use two frequency and time masks of size 30 and 40 respectively, and perform time warping with a warping factor of 5.

\textbf{Language Model:}
For decoding with a language model, a character-based 10-gram language model (LM) is trained using KenLM \cite{heafield2011kenlm}, and incorporated using shallow fusion.

\textbf{Hyperparameters:}
The model is trained using CTC loss \cite{Graves2006}. The Adam optimizer with learning scheduler given by \cite[Eq.~10]{8462506} is used with a learning rate factor of 5.0 and 25k training iterations for warmup. The models are trained for 100 epochs. The final model is obtained by averaging the 10 best models that have the lowest loss on the validation set. For inference, a beam search decoding algorithm is used with a beam size of 20. %

\section{Results}
\label{sec:results}
For all the experiments in this section, the baseline %
refers to the model that is trained on the labeled source domain training data from WSJ. Topline refers to models that are trained on the WSJ training data augmented with labeled data from the target domain.

\begin{figure}
    \centering
    \includegraphics[width=.82\linewidth]{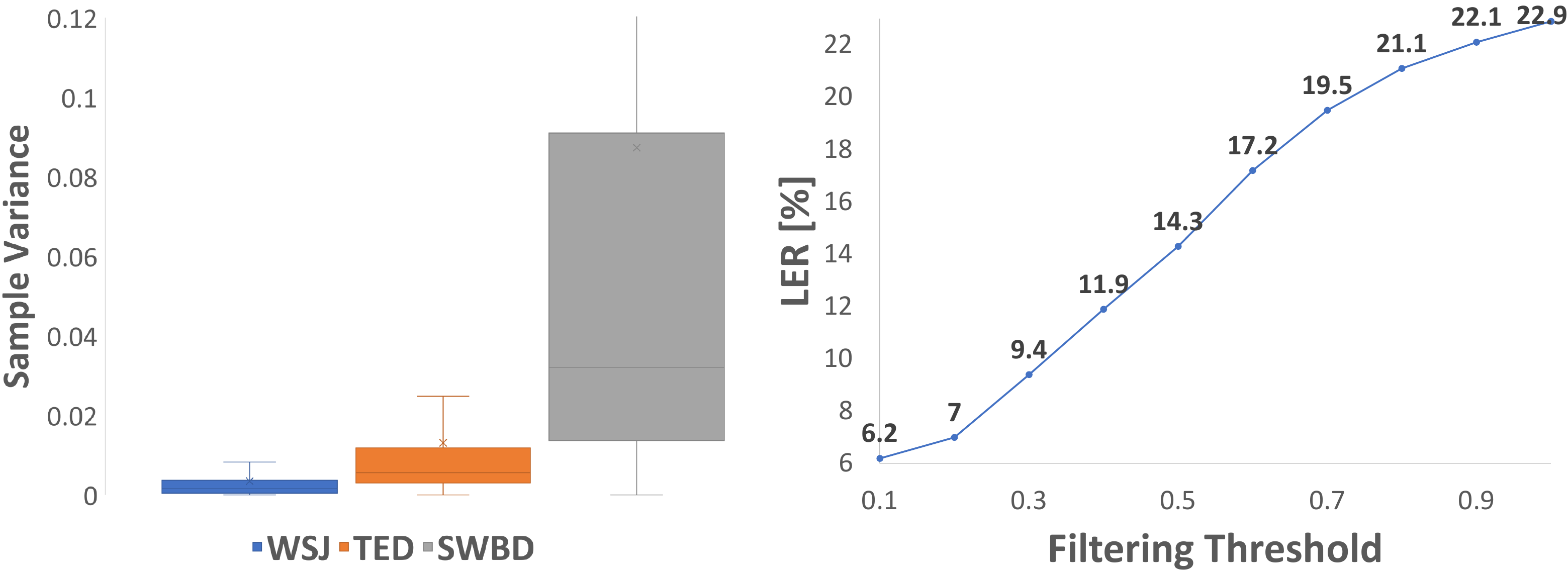}
    \caption{(Left) Distribution of the variance of the agreement between stochastic and deterministic samples as a measure of the model's uncertainty on source (WSJ) and target (TED, SWBD) test data. (Right) Influence of filtering threshold $\tau$ on LER [\%] of accepted pseudo-labeled utterances for TED.}\vspace{-.3cm}
    \label{fig:filtering_effect}
\end{figure}

In \textbf{Fig.~\ref{fig:filtering_effect}}, we show the efficacy of our filtering process in weeding out noisy pseudo-labels. The box plot shows the variance of the agreement (normalized edit distance) between the stochastic samples and the deterministic sample on source and target domain data points, as a measure of the model's prediction uncertainty. The source domain is WSJ and target domains are TED and SWBD. We train a base ASR model on the source training data and generate ten stochastic samples and a deterministic sample for each utterance in the test sets of the source and target domains via beam search with an LM trained on the source domain text. The variance of the edit distances is computed on each utterance. %
We see in the box plot that the model's uncertainty is significantly higher on target domain data than on source domain data, which concurs with our intuition. Furthermore, we show in the line graph the relationship between the filtering threshold $\tau$ and the label error rate (LER) on the pool $\PP$ of accepted pseudo-labeled utterances, for TED as the target domain. We see that at lower filtering thresholds, utterances in the set $\PP$ are much cleaner, as shown by the low LERs. %

\begin{table}[t]
    \caption{WSJ source to TED target domain adaptation using a 100k subset of unlabeled target domain data for self-training. LM trained on source domain text is used during pseudo-label generation} %
    \label{tab:wsj_ted_LM_100k}
    \centering
         \sisetup{table-format=2.1,round-mode=places,round-precision=1,table-number-alignment = center,detect-weight=true,detect-inline-weight=math}
    \resizebox{.99\linewidth}{!}{%
    \setlength{\tabcolsep}{2pt}
    \begin{tabular}{lS[table-format=3]SS[table-format=1.1]SSSS} \toprule
         & & \multicolumn{4}{c}{WER [\%]}& \multicolumn{2}{c}{WERR [\%]} \\
         \cmidrule(lr){3-6}\cmidrule(lr){7-8}
         Method & \multicolumn{1}{c}{$|\PP|$ [k]} & \multicolumn{1}{c}{$\PP$} &\multicolumn{1}{c}{\!\!WSJ/eval92\!}&\multicolumn{1}{c}{TED/dev}&\multicolumn{1}{c}{TED/test}&\multicolumn{1}{c}{WSJ/eval92}&\multicolumn{1}{c}{TED/test}\\
         \midrule
         \multicolumn{2}{l}{Baseline (w/o DataAug)} && 8.3 & 44.8&45.6&\-&\-\\
         Baseline &&& 6.8 & 37.1&35.0&0&0\\
         Topline & &&4.6 & 15.9&14.8&100&100\\
         \midrule
         \multicolumn{7}{l}{\textbf{\textit{First Self-Training Iteration}}}\\
         DUST1 ($\tau \!=\! 0.1$) &7&14.0&7.4&33.0&30.0&-27.2&24.7 \\
         DUST1 ($\tau \!=\! 0.3$) &38&25.0&\bfseries 6.1&\bfseries 30.0&\bfseries 26.8 &\bfseries 31.8&\bfseries 40.6\\
         DUST1 ($\tau \!=\! 0.5$) &70&34.0&6.3&31.3&27.6&22.7&36.6 \\
         DUST1 ($\tau \!=\! 0.7$) &90&39.0&6.2&31.1&27.9&27.2&35.1 \\
         ST1 (All) &100&42.0&6.3&31.0&27.7&22.7&36.1 \\
         \multicolumn{7}{l}{\textbf{\textit{Second Self-Training Iteration}}}\\
         DUST2 ($\tau \!=\! 0.1$) &35&14.8&6.9&30.1&27.3&-4.54&38.1 \\
         DUST2 ($\tau \!=\! 0.3$) &66&25.0&\bfseries 6.1&\bfseries 28.2&\bfseries 24.9& \bfseries 31.8&\bfseries 50.0\\
         DUST2 ($\tau \!=\! 0.5$) &88&33.8&6.6&29.4&25.6&9.10&46.5 \\
         DUST2 ($\tau \!=\! 0.7$) &95&39.0&6.7&29.3&26.1&4.54&44.1 \\
         ST2 (All) &100&42.0&6.3&29.4&25.8&22.7&45.5 \\
         \bottomrule
    \end{tabular}
    }\vspace{-.2cm}
\end{table}

In \textbf{Table~\ref{tab:wsj_ted_LM_100k}}, we first perform a comparison between classic ST, where all of the pseudo-labeled data is used, and DUST, where filtering is applied. We also investigate the effect of different thresholds on the downstream task performance. This set of experiments is performed using a 100k subset of unlabeled TED target domain data for self-training, and we use an LM trained on source domain text during the pseudo-label generation process via beam search.
We report both the word error rate (WER) and the WER recovery rate (WERR). The number $|\PP|$ of utterances in the filtered set $\PP$ and the WER on these utterances are also reported.
DUST with a filtering threshold of 0.3 performs better than ST1~(All), 26.8\% vs 27.7\%, while using approximately one-third of the pseudo-labeled data (38k vs 100k). When comparing the downstream task performance for different filtering thresholds, the following two observations can be made: %
first, the downstream task performance is substantially better than the baseline regardless of the filtering threshold;
second, best results are obtained when the filtering threshold $\tau$ is set to mid-range values, with the optimal setting being 0.3 based on the TED development set. 
From here on, we fix the threshold value $\tau$ to 0.3.
We next perform multiple iterations of self-training, %
where an LM trained on the source domain text is used for decoding during the pseudo-label generation process. The results are shown in \textbf{Table~\ref{tab:wsj_ted_LM_all_data}}. After each iteration, the number of selected utterances in $\PP$ increases, while the average WER on $\PP$ remains almost constant. The target domain WER shows clear improvements and DUST5 is able to recover 66\% of the WER compared to the topline, when using an LM trained on source domain text for decoding the test data, and an even slightly higher WERR of 67.4\% when using an LM trained on both source and target domain text instead.

\begin{table}[t]
    \caption{WSJ source to TED target domain adaptation. LM trained on source domain text is used during pseudo-label generation.} %
    \label{tab:wsj_ted_LM_all_data}
    \centering
     \sisetup{table-format=2.1,round-mode=places,round-precision=1,table-number-alignment = center,detect-weight=true,detect-inline-weight=math}
    \resizebox{.95\linewidth}{!}
    {%
    \setlength{\tabcolsep}{3pt}
    \begin{tabular}{lS[table-format=3]SSSSS} \toprule
         & & \multicolumn{3}{c}{WER [\%]}& \multicolumn{2}{c}{WERR [\%]} \\
         \cmidrule(lr){3-5}\cmidrule(lr){6-7}
         Method & \multicolumn{1}{c}{$|\PP|$ [k]} & \multicolumn{1}{c}{$\PP$} &\multicolumn{1}{c}{WSJ/eval92}&\multicolumn{1}{c}{TED/test}&\multicolumn{1}{c}{WSJ/eval92}&\multicolumn{1}{c}{TED/test}\\
         \midrule
         \multicolumn{7}{l}{\textbf{\textit{Decoding with an LM trained on source domain text}}} \\
         Baseline &&& 6.8 & 35.0&0&0\\
         ST1 (All) &265&41.1& 6.1 & 27.1&29.2&37.4\\
         DUST1 & 100&25.0&5.9 & 26.5&40.9&40.3\\
         DUST2 & 170&25.0&5.7 & 24.3&50.0&50.7\\
         DUST3 & 185&24.8&5.7 & 23.5&50.0&54.5\\
         DUST4 & 210&25.0&\bfseries 5.5 & 22.4&\bfseries 59.1&59.7\\
         \textbf{DUST5} & 230&25.4&5.6 & \bfseries 21.1&54.5&\bfseries 66.0\\
         Topline & &&4.4 & 13.9&100&100\\
         \midrule
         \multicolumn{7}{l}{\textbf{\textit{Decoding with an LM trained on source \& target domain text}}} \\
         Baseline & &&7.0 & 33.2&0&0 \\
         \textbf{DUST5} &230 &25.4&\bfseries 5.4 & \bfseries 19.3&\bfseries 59.2&\bfseries 67.4\\
         Topline & &&4.3 & 12.6&100&100\\
         \bottomrule
    \end{tabular}
    }\vspace{-.2cm}
\end{table}

In \textbf{Table~\ref{tab:wsj_ted_noLM}}, the impact of using an LM for the pseudo-label generation process is investigated. %
Our proposed filtering process generates multiple stochastic samples, which requires us to run beam search decoding multiple times with a different dropout initialization. This can be an expensive process, %
but it could be accelerated if the use of an LM for decoding during the pseudo-label generation process could be omitted.
The results in Table~\ref{tab:wsj_ted_noLM} as compared to Table~\ref{tab:wsj_ted_LM_all_data} show that in fact similar or better results can be achieved without the use of an LM for the pseudo-label generation process. First, here again DUST1 performs on par with classic self-training (ST1), while using only a fraction of pseudo-labeled utterances (see $|\PP|$ in Table 3).
Compared with the baseline, we are able to recover 80\% of the WERs when using no LM for decoding the test data, 79\% when using a source domain LM, and 76\% when using an LM that is trained on both source and target domain text.  %
In the latter case, we achieve a WER of $17.6\%$ on target domain using DUST, which is close to the topline WER of $12.6\%$, and outperforms the best WER of $19.3\%$ achieved in Table~\ref{tab:wsj_ted_LM_all_data}, where a source domain LM is used during the pseudo-label generation process. We suspect that using a source domain LM for the pseudo-label generation biases the generated pseudo-label towards the source domain text, hindering generalization, assuming the seed model is good enough without an LM to generate a sufficient amount of ``clean'' pseudo-labeled utterances that can be selected by DUST.

\begin{table}[t]
    \caption{WSJ source to TED target domain adaptation. No LM is used during pseudo-label generation.} %
    \label{tab:wsj_ted_noLM}
    \centering
     \sisetup{table-format=2.1,round-mode=places,round-precision=1,table-number-alignment = center,detect-weight=true,detect-inline-weight=math}
    \resizebox{.95\linewidth}{!}{%
    \setlength{\tabcolsep}{3pt}
    \begin{tabular}{lS[table-format=3]SSSSS} \toprule
         & & \multicolumn{3}{c}{WER [\%]}& \multicolumn{2}{c}{WERR [\%]} \\
         \cmidrule(lr){3-5}\cmidrule(lr){6-7}
         Method & \multicolumn{1}{c}{$|\PP|$ [k]} & \multicolumn{1}{c}{$\PP$} &\multicolumn{1}{c}{WSJ/eval92}&\multicolumn{1}{c}{TED/test}&\multicolumn{1}{c}{WSJ/eval92}&\multicolumn{1}{c}{TED/test}\\
         \midrule
         \multicolumn{7}{l}{\textbf{\textit{Decoding without LM}}}\\
         Baseline &&& 15.0 & 47.9&0&0\\
         ST1 (All) &265&53.4& 14.0 & 36.5&19.6&39.4 \\
         DUST1 &25&25.0&14.1&37.8&17.6&34.9 \\
         DUST2 &81&25.0&13.1&31.5&37.2&56.7\\
         DUST3 &136&25.8&13.1&28.1&37.2&68.5 \\
         DUST4 &167&25.0&\bfseries 12.5 &25.8&\bfseries 49.0&76.4 \\
         \textbf{DUST5} &178&23.8&12.6&\bfseries 24.7 &47.0&\bfseries 80.2 \\
         Topline & &&9.9 & 19.0&100&100\\
         \midrule
         \multicolumn{7}{l}{\textbf{\textit{Decoding with an LM trained on source domain text}}}\\
         Baseline & &&6.8 & 35.0&0&0\\
         ST1 (All) &265&53.4& 6.3 & 26.8&20.8&38.8 \\
         DUST1 &25&25.0&6.1&27.9&29.2&33.6 \\
         DUST2 &81&25.0&6.1&23.3&29.2&55.4\\
         DUST3 &136&25.8&\bfseries 5.6&20.9&\bfseries 50.0&66.8 \\
         DUST4 &167&25.0& 6.0 &19.4&33.3&73.9\\
         \textbf{DUST5} & 178&23.8&5.7 &\bfseries 18.4&45.8&\bfseries 78.6 \\
         Topline & &&4.4 & 13.9&100&100 \\
         \midrule
         \multicolumn{7}{l}{\textbf{\textit{Decoding with an LM trained on source \& target domain text}}} \\
         Baseline & &&7.0 & 33.2&0&0 \\
         \textbf{DUST5} & 178&23.8&\bfseries 5.6 &\bfseries  17.6&\bfseries 51.8&\bfseries 75.7\\
         Topline & &&4.3 & 12.6&100& 100\\
         \bottomrule
    \end{tabular}
    }\vspace{-.1cm}
\end{table}

In \textbf{Table 4}, the results for SWBD as the target domain are presented. The domain mismatch, as evident from the baseline results, is quite severe. Nevertheless, DUST is able to improve over the baseline by $22.4$ percentage points (pp) and is able to recover 65\% of WER when using a source domain LM for decoding the test data, and 61\% when using an LM trained on both source and target domain text.
For SWBD, we were not able to obtain good performance without using an LM during the filtering process in the early DUST iterations, probably due to the severity of domain mismatch. However, we observe that by removing the LM from the pseudo-label generation process after a few DUST iterations, better results can be obtained compared to using an LM throughout (see ``DUST4,5 (w/o LM)'' vs ``DUST4,5''). Future work should explore when to remove the LM. While the results are encouraging, there is still a large gap between DUST and the topline, which could be addressed by relaxing one of our assumptions regarding not having access to any target domain text data. We leave this investigation for future work. Importantly, the results shown in Table~\ref{tab:wsj_swbd} demonstrate that DUST1 substantially outperforms ST1, which shows that pseudo-label filtering is essential under severe domain mismatched scenarios.

Finally, in \textbf{Table 5}, we investigate whether DUST could be effectively combined %
with a self-supervised representation learning approach %
for low-resource speech recognition. We train the base source model on just 3 hours of labeled source data (WSJ) using Wav2Vec (W2V) \cite{Baevski2020wav2vec} features as input to the model to remove some of the domain mismatch compared to a baseline trained without Wav2Vec. %
Four DUST iterations using the unlabeled source (WSJ) and target domain data (TED) significantly improve the performance: DUST improves over the baseline by 32.2 pp and 51.0 pp for the source and target domains, recovering 83\% and 62\% of the WER, respectively, while Wav2Vec alone only improves by 6.2 pp and 17.4 pp. Similarly to what was observed in Table 4, removing the source LM from the pseudo-label generation process after a couple of DUST iterations leads to better results. W2V is trained on Librispeech and the features are only used to train the base model. %

\begin{table}[t]
    \caption{WSJ source to SWBD target domain adaptation.} %
    \label{tab:wsj_swbd}
    \centering
     \sisetup{table-format=2.1,round-mode=places,round-precision=1,table-number-alignment = center,detect-weight=true,detect-inline-weight=math}
    \resizebox{.999\linewidth}{!}{%
    \setlength{\tabcolsep}{3pt}
    \begin{tabular}{lS[table-format=3]SSSSS} \toprule
         & & \multicolumn{3}{c}{WER [\%]}& \multicolumn{1}{c}{WERR [\%]} \\
         \cmidrule(lr){3-5}\cmidrule(lr){6-6}
         Method & \multicolumn{1}{c}{$|\PP|$ [k]} & \multicolumn{1}{c}{$\PP$} &\multicolumn{1}{c}{WSJ/eval92}&\multicolumn{1}{c}{SWBD/eval2000}&\multicolumn{1}{c}{SWBD/eval2000}\\
         \midrule
         \multicolumn{6}{l}{\textbf{\textit{Decoding with an LM trained on source domain text}}}\\
         \multicolumn{2}{l}{Baseline (w/o DataAug)} & &8.3&82.3&&\\
         Baseline & &&6.8 & 64.1&0\\
         ST1 (All) &192&63.7&7.1&56.8&19.2\\
         DUST1 & 7&33& 7.6& 50.0&37.1 \\
         DUST2 & 30&35& 7.3& 47.3&44.2 \\
         DUST3 & 68&35& 6.9& 44.1&52.6 \\
         DUST4 & 108&34.9& 7.0& 42.7&56.3 \\
         DUST5 & 125&35.4&7.1& 41.7&58.9 \\
        DUST4 (w/o LM)\!\! & 65&32.6&6.7& 40.3& 62.6\\
         \textbf{DUST5 (w/o LM)\!\!} &78&31.7&\bfseries 6.7&\bfseries 39.2& \bfseries 65.5\\
         Topline & && 6.6& 26.1&100 \\
         \midrule
         \multicolumn{6}{l}{\textbf{\textit{Decoding with an LM trained on source \& target domain text}}} \\
         Baseline & && 7.2& 61.7&0 \\
         DUST5 & 125&35.4& 7.1& 39.9& 56.2\\
         \textbf{DUST5 (w/o LM)\!\!} & 78&31.2&\bfseries  6.7& \bfseries 38.1&\bfseries 60.8 \\
         Topline & && 6.6& 22.9& 100\\
         \bottomrule
    \end{tabular}
    }\vspace{-.1cm}
\end{table}

\begin{table}
    \caption{DUST for low-resource ASR. Decoding with source LM} %
    \label{tab:wav2vec}
    \centering
     \sisetup{table-format=2.1,round-mode=places,round-precision=1,table-number-alignment = center,detect-weight=true,detect-inline-weight=math}
    \resizebox{.999\linewidth}{!}{%
    \setlength{\tabcolsep}{3pt}
    \begin{tabular}{lS[table-format=3]SSSSS} \toprule
         & & \multicolumn{3}{c}{WER [\%]}& \multicolumn{2}{c}{WERR [\%]} \\
         \cmidrule(lr){3-5}\cmidrule(lr){6-7}
         Method & \multicolumn{1}{c}{$|\PP|$ [k]} & \multicolumn{1}{c}{$\PP$} &\multicolumn{1}{c}{WSJ/eval92}&\multicolumn{1}{c}{TED/test}&\multicolumn{1}{c}{WSJ/eval92}&\multicolumn{1}{c}{TED/test}\\
         \midrule
         Baseline &&& 43.2 & 95.6&0&0\\
         Wav2Vec \cite{Baevski2020wav2vec} &&& 37.0 & 78.2&16.0&21.3\\
         DUST1 & 60&39.1&16.0 & 60.0&70.1&43.6\\
         DUST2 & 112&38.5& 12.2 & 54.4&79.9&50.4 \\
         DUST3 &171&42.6&11.0&50.2&82.5&53.1\\
         DUST3 (w/o LM)\!\! &94&35.4&\bfseries 10.8 & 48.1&\bfseries 83.5&58.1\\
         \textbf{DUST4 (w/o LM)\!\!}&141&38.8&11.0& \bfseries 44.6& 82.9&\bfseries 62.4\\
         Topline & &&4.4 & 13.9&100&100\\
         \bottomrule
    \end{tabular}
    }\vspace{-.3cm}
\end{table}

\section{Conclusion}
In this work, we proposed DUST, a dropout-based uncertainty-driven self-training method for unsupervised domain adaptation. 
Through several experiments transferring from WSJ to TED-LIUM~3 and SWITCHBOARD, we show that DUST significantly improves performance over the baseline model trained only on the labeled source domain data, and is able to recover 60\% to 80\% of the WER on the target domain by using only unlabeled speech from that domain. Future work includes performing a thorough investigation of the combination of DUST with self-supervised representation learning approaches for low-resource semi-supervised speech recognition and domain adaptation, and %
exploring the combination of DUST with other complementary semi-supervised methods for E2E ASR such as the recently proposed Graph-based Temporal Classification (GTC) \cite{moritz2020gtc}.

\vfill\pagebreak

\bibliographystyle{IEEEtran}
\bibliography{refs}

\end{document}